\documentclass[sigconf]{aamas} 

\usepackage{balance} 
\usepackage{multirow} 
\usepackage{makecell} 
\usepackage{algorithm} 
\usepackage{algpseudocode}

\makeatletter
\gdef\@copyrightpermission{
  \begin{minipage}{0.2\columnwidth}
   \href{https://creativecommons.org/licenses/by/4.0/}{\includegraphics[width=0.90\textwidth]{by}}
  \end{minipage}\hfill
  \begin{minipage}{0.8\columnwidth}
   \href{https://creativecommons.org/licenses/by/4.0/}{This work is licensed under a Creative Commons Attribution International 4.0 License.}
  \end{minipage}
  \vspace{5pt}
}
\makeatother

\setcopyright{ifaamas}
\acmConference[AAMAS '26]{Proc.\@ of the 25th International Conference
on Autonomous Agents and Multiagent Systems (AAMAS 2026)}{May 25 -- 29, 2026}
{Paphos, Cyprus}{C.~Amato, L.~Dennis, V.~Mascardi, J.~Thangarajah (eds.)}
\copyrightyear{2026}
\acmYear{2026}
\acmDOI{}
\acmPrice{}
\acmISBN{}

\acmSubmissionID{<<submission id>>}

\title[BVME]{Bandwidth-constrained Variational Message Encoding for Cooperative Multi-agent Reinforcement Learning}

\author{Wei Duan}
\affiliation{
  \institution{University of Technology Sydney}
  \city{Sydney}
  \country{Australia}}
\email{wei.duan@uts.edu.au}

\author{Jie Lu}
\affiliation{
  \institution{University of Technology Sydney}
  \city{Sydney}
  \country{Australia}}
\email{jie.lu@uts.edu.au}

\author{En Yu}
\affiliation{
  \institution{University of Technology Sydney}
  \city{Sydney}
  \country{Australia}}
\email{en.yu-1@uts.edu.au}

\author{Junyu Xuan}
\affiliation{
  \institution{University of Technology Sydney}
  \city{Sydney}
  \country{Australia}}
\email{junyu.xuan@uts.edu.au}

\begin{abstract}
Graph-based multi-agent reinforcement learning (MARL) enables coordinated behavior under partial observability by modeling agents as nodes and communication links as edges. 
While recent methods excel at learning sparse coordination graphs—determining \emph{who} communicates with \emph{whom}—they do not address \emph{what} information should be transmitted under hard bandwidth constraints. 
We study this bandwidth-limited regime and show that naïve dimensionality reduction consistently degrades coordination performance.
Hard bandwidth constraints force selective encoding, but deterministic projections lack mechanisms to control how compression occurs.
We introduce \textbf{Bandwidth-constrained Variational Message Encoding (BVME)}, a lightweight module that treats messages as samples from learned Gaussian posteriors regularized via KL divergence to an uninformative prior. 
BVME's variational framework provides principled, tunable control over compression strength through interpretable hyperparameters, directly constraining the representations used for decision-making.
Across SMACv1, SMACv2, and MPE benchmarks, BVME achieves comparable or superior performance while using 67--83\% fewer message dimensions, with gains most pronounced on sparse graphs where message quality critically impacts coordination.
Ablations reveal U-shaped sensitivity to bandwidth, with BVME excelling at extreme ratios while adding minimal overhead.
\end{abstract}


\newcommand{\BibTeX}{\rm B\kern-.05em{\sc i\kern-.025em b}\kern-.08em\TeX}

\begin{document}


\pagestyle{fancy}
\fancyhead{}


\maketitle 

\section{Introduction}

Cooperative multi-agent reinforcement learning (MARL) under partial observability fundamentally requires information sharing among agents~\cite{foerster2016learning,liu2023partially,Duan2025bayesian,DBLP:conf/ifaamas/VarelaSM25,yang2025adapting,yang2025resilientcontrastivepretrainingnonstationary}. Graph-based formulations, where agents are nodes and edges represent coordination dependencies, have become a dominant framework for scalable cooperation~\cite{boehmer2020deep,jiang2020graph,DBLP:conf/atal/LiGMAK21,DBLP:conf/atal/WeilBAM24,yu2025learning,yu2025drift,yang2025walking}. When combined with graph neural networks (GNNs)~\cite{DBLP:conf/iclr/KipfW17,10255371,DBLP:conf/iske/DuanX021,DBLP:conf/aaai/DuanXQ022,duan2024layerdiverse,DBLP:journals/eswa/YaoHLDQYS25}, these approaches enable end-to-end learning of coordination policies through differentiable message passing, achieving strong performance in domains ranging from cooperative navigation to strategy games~\cite{he2024commformer,wang2024magec,DBLP:conf/atal/WeilBAM24}. Despite this success, a fundamental challenge remains underexplored: how to optimally encode information when communication capacity is limited.

\begin{figure}[t]
\centering
\includegraphics[width=\columnwidth]{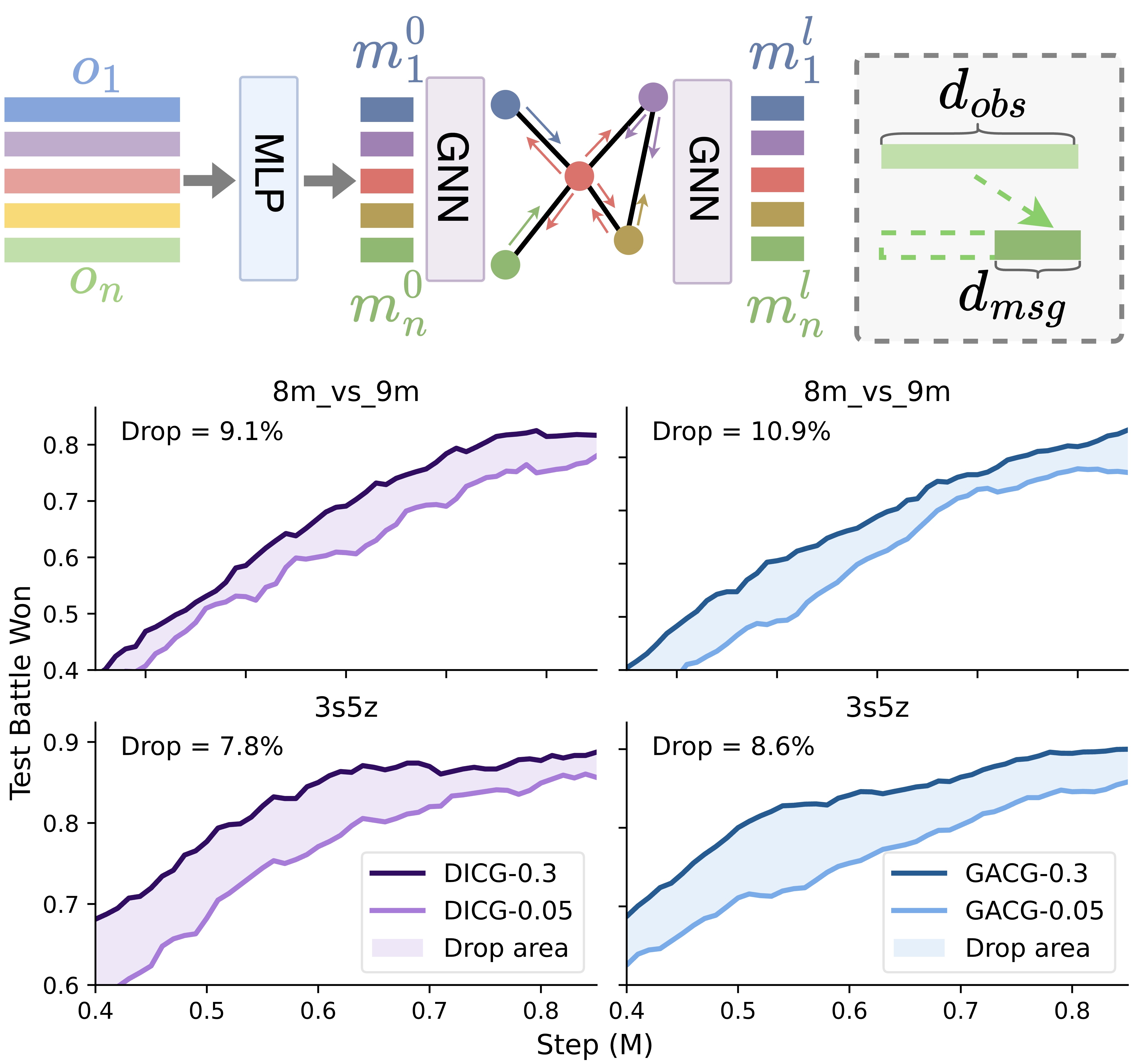}
\caption{
\textbf{Bandwidth constraints degrade performance.}
\emph{Top:} Standard GNN-based MARL encodes observations $o_i \!\in\! \mathbb{R}^{d_{\mathrm{obs}}}$ into messages $m_i^l \!\in\! \mathbb{R}^{d_{\mathrm{msg}}}$. Compression ratio $r = d_{\mathrm{msg}}/d_{\mathrm{obs}}$ controls bandwidth.
\emph{Bottom:} Test win rates on SMACv1 when reducing $r$ from 0.3 to 0.05. Shaded regions show performance loss (drop area). Sparse graphs (GACG) are more sensitive to bandwidth limits than dense graphs (DICG).
}
\label{fig:bandwidth-drop}
\end{figure}

Existing work addresses two complementary aspects of multi-agent communication. \textbf{Topology-focused methods} learn \textbf{who} should communicate: attention mechanisms~\cite{jiang2018atoc,iqbal2019actor,kim2019schednet,DBLP:conf/aaai/LiuWHHC020,DBLP:conf/atal/LiGMAK21,shao2023complementary} dynamically weight interactions, while sparse coordination graphs~\cite{DBLP:conf/ijcai/00030X24,DuanLTS-CG2024} prune unnecessary connections. These approaches implicitly assume that once a link is established, sufficient bandwidth exists to transmit high-dimensional features. In contrast, \textbf{content-focused methods}~\cite{sukhbaatar2016commnet,foerster2016learning,DBLP:conf/iclr/SinghJS19,das2019tarmac,DBLP:conf/iclr/0001WZZ20} address \textbf{when} and \textbf{what} to communicate through gating and information-theoretic regularization, but typically omit explicit graph structure. Recent work has explored multi-level communication~\cite{zhang2024seqcomm} and interpretable information gating~\cite{icml2025infogating}, while MAGI~\cite{DingMAGI2024} integrates information bottlenecks into GNNs layers to filter noisy channels. However, these approaches share a common limitation: they do not explicitly model hard bandwidth limits—scenarios where message dimensionality is strictly bounded (e.g., 5–10\% of observation size) due to physical or scalability constraints. MAGI~\cite{DingMAGI2024}, for instance, relies on dense attention mechanisms rather than learned sparse graphs, while information-theoretic methods typically regulate soft capacity rather than enforcing dimensional constraints.

As illustrated in Fig.~\ref{fig:bandwidth-drop} (top), standard GNN-based MARL encodes each agent's observation $o_i \!\in\! \mathbb{R}^{d_{\mathrm{obs}}}$ through an MLP into an initial message $m_i^0$, which is refined via message-passing layers into $m_i^l \!\in\! \mathbb{R}^{d_{\mathrm{msg}}}$. The compression ratio $r = d_{\mathrm{msg}}/d_{\mathrm{obs}}$ defines communication bandwidth. Existing methods~\cite{DBLP:conf/atal/LiGMAK21,DBLP:conf/ijcai/00030X24,DuanLTS-CG2024} apply deterministic linear projections to satisfy dimensional constraints, but when $r$ is small ($\le 0.1$), this approach discards information indiscriminately. To quantify this effect, we conduct a controlled study on DICG~\cite{DBLP:conf/atal/LiGMAK21} (dense) and GACG~\cite{DBLP:conf/ijcai/00030X24} (sparse) using SMACv1~\cite{DBLP:conf/atal/SamvelyanRWFNRH19}, progressively reducing $r$ from 0.3 to 0.05. 
Both architectures suffer consistent performance drops, with sparse graphs exhibiting sharper degradation (Fig.~\ref{fig:bandwidth-drop}, bottom). 
We quantify this via \emph{drop area}—the cumulative loss between bandwidth curves. 
Sparse graphs are more sensitive because each edge must carry more task-critical information, demonstrating that topology and bandwidth interact in task-dependent ways. 

Such bandwidth constraints arise naturally in real-world deployments, such as warehouse robot swarms that train centrally using full state observations but execute via local wireless radios with limited throughput.
In these scenarios, hard bandwidth constraints force selective encoding, yet deterministic projections lack mechanisms to control how compression occurs—they discard information uniformly regardless of task relevance.
This motivates our central question: \textbf{How can agents achieve effective coordination when communication bandwidth is severely limited?}

To address this question, we propose \textbf{Bandwidth-constrained Variational Message Encoding (BVME)}.
BVME models each agent's message as a sample from a learned Gaussian distribution, regularized toward an uninformative prior via Kullback-Leibler (KL) divergence.
This variational framework provides principled, tunable control over compression through interpretable hyperparameters, enabling: \textbf{(1)} direct regulation of how much information agents transmit via the KL penalty; \textbf{(2)} end-to-end learning of task-relevant representations under bandwidth constraints; \textbf{(3)} flexible adjustment of compression strength without architecture changes. A critical design element is \emph{on-path} coupling: BVME directly injects sampled messages into each agent's $Q$-network, ensuring the KL regularization operates on the same representations that drive coordination decisions, rather than filtering auxiliary signals that bypass decision-making.

We instantiate BVME on GACG~\cite{DBLP:conf/ijcai/00030X24}, which learns sparse relations between agents and groups, demonstrating how variational encoding complements structured topologies.
We choose \textbf{SMACv1}~\cite{DBLP:conf/atal/SamvelyanRWFNRH19}, \textbf{SMACv2}, and \textbf{MPE-Tag} to evaluate our method, comparing against QMIX~\cite{DBLP:conf/icml/RashidSWFFW18}, DICG~\cite{DBLP:conf/atal/LiGMAK21}, and GACG~\cite{DBLP:conf/ijcai/00030X24}.
Under severe bandwidth constraints (5\%), BVME achieves higher win rates and faster convergence.
Ablations reveal a U-shaped performance curve: substantial gains at low bandwidth, marginal benefits at moderate compression, and neutral impact at high capacity.
On-path coupling proves essential, substantially outperforming off-path regularization.

\noindent \textbf{Our contributions are:}

\noindent \textbf{(1)} We propose BVME, a variational framework providing \textbf{principled, tunable control over message compression} through KL regularization. Unlike deterministic projections that discard information uniformly, BVME enables explicit capacity control via interpretable hyperparameters ($r$, $\sigma_0$, $\lambda_{\mathrm{KL}}$).

\noindent \textbf{(2)} We identify a U-shaped sensitivity to bandwidth: BVME excels at extreme compression ($r \leq 0.05$) where it filters noise and prioritizes coordination-critical features, with on-path coupling essential for effectiveness.

\noindent \textbf{(3)} Experiments across SMACv1/v2 and MPE-Tag show BVME enables stable learning where baselines fail (5\% bandwidth) and achieves comparable performance with 67–83\% fewer message dimensions on sparse graphs.

\begin{figure*}[t]
\centering
\includegraphics[width=\textwidth]{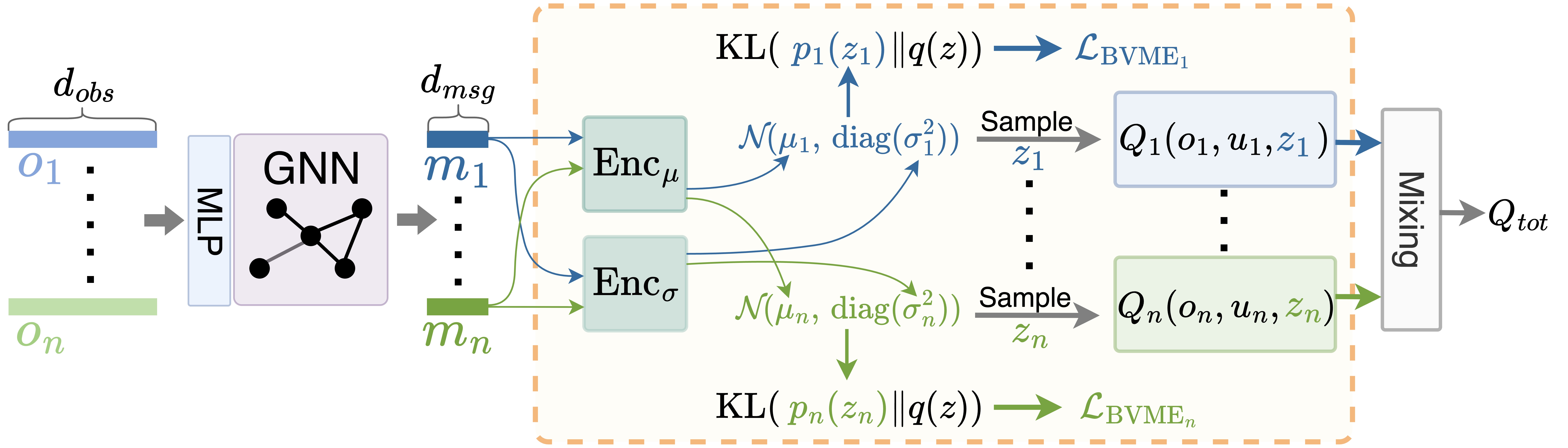}
\caption{\textbf{BVME architecture.} 
\emph{Standard approach (left):} GNN-based MARL encodes observations $\{o_i\}$ via MLP compression followed by graph convolution into messages $\{m_i\}$ that directly feed $Q$-functions. 
\emph{BVME (right, orange box):} Each $m_i$ parameterizes a Gaussian posterior via encoders $\mathrm{Enc}_\mu$ and $\mathrm{Enc}_\sigma$. 
Sampled messages $z_i$ are used for $Q_i$ estimation, while KL divergence to a prior $q(z)=\mathcal{N}(0,\sigma_0^2 I)$ enforces bandwidth constraints. 
The compression ratio $r = d_{\mathrm{msg}}/d_{\mathrm{obs}}$ controls message dimensionality.
Training combines TD loss with BVME regularization.}
\label{fig:flowchart}
\end{figure*}

\section{Related Work}
\subsection{Graph-structured coordination in MARL}

Graph neural networks have become a dominant framework for cooperative MARL under partial observability~\cite{boehmer2020deep,jiang2020graph}, often combined with value decomposition~\cite{DBLP:conf/atal/SunehagLGCZJLSL18,DBLP:conf/icml/RashidSWFFW18,DBLP:conf/nips/PhanRBAGL21} for decentralized execution with centralized training. Methods for learning \emph{who} communicates can be categorized by their induced communication topology.

\textbf{Dense communication.} 
DCG~\cite{boehmer2020deep} and DICG~\cite{DBLP:conf/atal/LiGMAK21} use complete coordination graphs or dense neural networks to model all agent interactions, while attention-based approaches like ATOC~\cite{jiang2018atoc} and MAAC~\cite{iqbal2019actor} apply learned weights across the full team. Though adaptively weighted, these methods scale poorly as team size grows. \textbf{Sparse coordination graphs.} 
SOP-CG~\cite{DBLP:conf/icml/YangDRW0Z22}, GACG~\cite{DBLP:conf/ijcai/00030X24}, LTS-CG~\cite{DuanLTS-CG2024}, and CommFormer~\cite{he2024commformer} learn sparse topologies through value decomposition over restricted structures, group-aware distributions, temporal inference, or bi-level optimization, enabling scalable coordination by restricting communication to task-relevant subsets.

Both paradigms optimize \emph{who} communicates but treat messages as deterministic embeddings, overlooking \emph{what} information should be transmitted under bandwidth constraints. When message dimensionality is limited, even well-structured topologies may propagate redundant or noisy content.

\subsection{Communication-efficient MARL}
Complementary research explores \emph{when} and \emph{what} to communicate, often without explicit graph structure. 
IC3Net~\cite{DBLP:conf/iclr/SinghJS19} and TarMAC~\cite{das2019tarmac} learn dynamic communication policies, while information-theoretic methods like NDQ~\cite{DBLP:conf/iclr/0001WZZ20} promote task-relevant messaging by penalizing unnecessary entropy. 
Recent work explores hierarchical~\cite{zhang2024seqcomm} and interpretable~\cite{icml2025infogating} communication mechanisms. 
Liu \& Bai~\cite{liu2023partially} provide theoretical analysis of information sharing benefits under partial observability.
However, these methods typically lack structural biases that enhance scalability and credit assignment.

\noindent\textbf{Prior work on bandwidth-limited MARL} addresses orthogonal aspects of communication efficiency.
Scheduling methods~\cite{DBLP:conf/aaai/MaoZXGN20,kim2019schednet} learn \emph{when/who} communicates via gating or Top-$k$ selection, reducing \emph{frequency} without compressing \emph{content}.
MAGI~\cite{DingMAGI2024} applies information bottlenecks to GNN stacks but targets adversarial robustness via dense attention rather than dimensional constraints with sparse graphs.
Existing methods treat bandwidth heuristically through projections or auxiliary losses—without explicit control over the message dimensionality that directly informs decisions.
BVME addresses this via on-path KL regularization, ensuring bandwidth limits constrain representations that drive coordination.

\section{Preliminaries}

\subsection{Cooperative Dec-POMDPs}
Cooperative multi-agent tasks can be modeled as a Decentralized Partially Observable Markov Decision Process (Dec-POMDP)~\cite{DBLP:series/sbis/OliehoekA16} with tuple
$\langle \mathcal{A}, \mathcal{S}, \{\mathcal{U}_i\}_{i=1}^n, P, \{\mathcal{O}_i\}_{i=1}^n, \{\pi_i\}_{i=1}^n, R, \gamma \rangle$.
At time $t$, each agent $i \in \mathcal{A}=\{1,\dots,n\}$ receives a local observation $o_i^t \in \mathcal{O}_i$, maintains an action-observation history $\tau_i^t=(o_i^0,u_i^0,\dots,o_i^t)$, and selects an action $u_i^t \in \mathcal{U}_i$ via a stochastic policy $\pi_i(u_i^t \mid \tau_i^t)$.
The joint action $\mathbf{u}^t=(u_1^t,\dots,u_n^t)$ induces the next state $s^{t+1}\!\sim\!P(\cdot \mid s^t,\mathbf{u}^t)$ and yields a shared reward $r^t=R(s^t,\mathbf{u}^t)$.
The objective is to maximize the expected discounted return:
\begin{equation}
Q_{\mathrm{tot}}(s,\mathbf{u})
\;=\;
\mathbb{E}\!\left[\sum_{t=0}^{\infty}\gamma^t r^t \;\middle|\; s^0\!=\!s,\, \mathbf{u}^0\!=\!\mathbf{u}\right],
\qquad \gamma \in [0,1).
\end{equation}

We adopt the CTDE paradigm~\cite{DBLP:conf/icml/RashidSWFFW18,DBLP:conf/nips/YuVVGWBW22}: during training, a central learner accesses global state $s^t$ and all observations $\{o_i^t\}_{i=1}^n$ to optimize policies and coordination structures; during execution, each agent independently selects actions via $\pi_i(\tau_i^t)$ using only local history and received messages.
Bandwidth constraints $r = d_{\mathrm{msg}}/d_{\mathrm{obs}}$ on inter-agent communication ensure deployability on resource-constrained platforms.
A widely used approach is \emph{value decomposition}, where per-agent $Q_i(\tau_i, u_i)$ are combined to calculate $Q_{\mathrm{tot}}$.

\subsection{Graph Neural Networks for Agent Communication}
Graph neural networks (GNNs) are widely used in cooperative MARL to support inter-agent communication via a \emph{coordination graph} whose nodes are agents and edges represent communication links.
Given an adjacency matrix $\widetilde{A} \in \mathbb{R}^{n \times n}$ and agent features $X \in \mathbb{R}^{n \times d}$, a message-passing layer computes
\begin{equation}
M \;=\; \phi\!\big(\widetilde{A}\, X\, W\big),
\end{equation}
where $W$ is a learnable weight matrix and $\phi(\cdot)$ is a nonlinearity. After $L$ layers, the final messages $M^{(L)}$ are integrated into per-agent utilities:
\begin{equation}
Q_i(\tau_i, u_i) \;=\; f_{\theta_i}\!\big(o_i^t, u_i, m_i^{(L)}\big),
\end{equation}
where $m_i^{(L)}$ is the $i$-th row of $M^{(L)}$. The utilities $\{Q_i\}_{i=1}^n$ are then combined via value decomposition to produce $Q_{\mathrm{tot}}$.

\subsection{Group-Aware Coordination Graphs (GACG)}
\label{sec:prelim_gacg}

A key challenge in GNN-based MARL is determining the adjacency $\widetilde{A}$ that governs message passing.
We build on GACG~\cite{DBLP:conf/ijcai/00030X24}, which learns sparse, group-aware coordination graphs by representing the adjacency as a Gaussian distribution:
\begin{equation}
\mathrm{vec}(\widetilde{A}^t) \sim \mathcal{N}\!\big(\mu_A^t, \Sigma_A^t\big),
\label{eq:gacg_sampling}
\end{equation}
where $\mu_A^t$ encodes pairwise agent importance derived from current observations, and $\Sigma_A^t$ captures group-level correlations estimated from recent trajectory history. This formulation enables both sparse edge selection and discovery of coordinated sub-teams.

GACG is trained with a QMIX-style TD loss and a group regularizer that encourages within-group cohesion and between-group specialization:
\begin{equation}
\label{eq:gacg_obj}
\mathcal{L}_{\text{GACG}} \;=\; \mathcal{L}_{\text{TD}} \;+\; \lambda_g\,\mathcal{L}_g,
\end{equation}
where $\mathcal{L}_g$ compares intra-group policy similarity to inter-group diversity (see~\cite{DBLP:conf/ijcai/00030X24} for details).
In our work, we adopt GACG's graph learning mechanism to obtain $\widetilde{A}^t$, and focus on optimizing the message content $M$ that flows on this learned sparse topology.

While GACG optimizes \emph{which} edges to activate, it treats the messages $M$ transmitted on these edges as deterministic embeddings. In the next section, we introduce BVME, which addresses \emph{what} information should be encoded into $M$ when message dimensionality is constrained.

\section{Method: Bandwidth-constrained Variational Message Encoding}
\label{sec:method}
Graph-structured coordination in MARL typically exchanges full or lightly projected messages between agents.
When bandwidth is tight ($d_{\mathrm{msg}} \ll d_{\mathrm{obs}}$), naïve linear projection degrades coordination performance (Fig.~\ref{fig:bandwidth-drop}).
We propose \emph{Bandwidth-constrained Variational Message Encoding} (BVME), which treats messages as samples from learned Gaussian posteriors regularized via KL divergence~\cite{DBLP:journals/corr/KingmaW13}.
BVME provides principled, tunable control over compression strength through: compression ratio $r = d_{\mathrm{msg}}/d_{\mathrm{obs}}$, prior scale $\sigma_0$, and KL weight $\lambda_{\mathrm{KL}}$, enabling agents to learn compact representations under strict bandwidth limits.

\subsection{Standard Message Passing}
\label{sec:standard_mp}
Given a learned adjacency $\widetilde{A}$ (Sec.~\ref{sec:prelim_gacg}) and agent features $X^{(l-1)} \in \mathbb{R}^{n \times d_{l-1}}$, 
a standard GNN message-passing layer computes:
\begin{equation}
M^{(l)} \;=\; \phi_l\!\big(\widetilde{A}\, X^{(l-1)} W_l\big),
\label{eq:gnn_mp}
\end{equation}
where $W_l \in \mathbb{R}^{d_{l-1} \times d_l}$ is a learnable weight matrix and $\phi_l$ is a nonlinearity.
To satisfy a bandwidth budget $r = d_{\mathrm{msg}}/d_{\mathrm{obs}}$, 
one sets the final layer width $d_L = d_{\mathrm{msg}}$. 
This projection respects the dimensional constraint but is task-agnostic: when $r \le 0.3$, performance drops sharply (Fig.~\ref{fig:bandwidth-drop}).

\subsection{Variational Message Encoding}
\label{sec:variational_encoding}

\textbf{From Hard Constraints to Tunable Compression Control.}
Hard bandwidth constraints force selective encoding, but deterministic projections lack mechanisms to control how compression occurs—they discard information uniformly regardless of task relevance.
By sampling messages from learned distributions regularized via KL divergence to an uninformative prior  (Sec.~\ref{sec:kl_penalty}), the KL penalty directly controls how much information each agent transmits. \textbf{Crucially, agents can adapt compression strength through tunable hyperparameters ($\sigma_0$, $\lambda_{\mathrm{KL}}$), unlike fixed projections that compress uniformly regardless of coordination needs.}
This enables: \textbf{(1)} explicit capacity control via tunable hyperparameters ($\sigma_0$, $\lambda_{\mathrm{KL}}$); \textbf{(2)} end-to-end learning of task-relevant representations under bandwidth constraints; \textbf{(3)} stochastic regularization that may improve generalization.

\paragraph{\textbf{Gaussian parameterization.}} 
Rather than directly using the GNN outputs $M^{(l)} = \{m_1^{(l)}, \dots, m_n^{(l)}\}$ from Eq.~\eqref{eq:gnn_mp}, we use each $m_i^{(l)} \in \mathbb{R}^{d_{\mathrm{msg}}}$ to parameterize a diagonal Gaussian posterior via lightweight encoders:
\begin{align}
\mu_i^{(l)} &= \mathrm{Enc}_\mu\!\big(m_i^{(l)}\big), \qquad
\log \sigma_i^{2(l)} \;=\; \mathrm{Enc}_\sigma\!\big(m_i^{(l)}\big),
\end{align}
where $\mathrm{Enc}_\mu, \mathrm{Enc}_\sigma : \mathbb{R}^{d_{\mathrm{msg}}} \to \mathbb{R}^{d_{\mathrm{msg}}}$ are single-layer MLPs.
These encoders map each GNN message to the mean and log-variance of a distribution, enabling the model to learn which message features are certain (low variance) versus uncertain (high variance).
Agent $i$'s message distribution is then:
\begin{equation}
p_i^{(l)}(z) = \mathcal{N}\!\big(z \mid \mu_i^{(l)},\, \mathrm{diag}(\sigma_i^{2(l)})\big).
\end{equation}
Each agent samples its Bandwidth-constrained message via reparameterization:
\begin{equation}
z_i^{(l)} = \mu_i^{(l)} + \sigma_i^{(l)} \odot \varepsilon, 
\quad \varepsilon \sim \mathcal{N}(0,I),
\label{eq:reparam}
\end{equation}
where $\odot$ denotes element-wise multiplication.
This standard reparameterization trick allows gradients to flow through the stochastic sampling operation.
Stacking sampled messages across agents yields:
\begin{equation}
Z^{(l)} = [z_1^{(l)},\dots,z_n^{(l)}]^\top \in \mathbb{R}^{n \times d_{\mathrm{msg}}},
\end{equation}
which replaces $M^{(l)}$ in subsequent GNN layers or as input to $Q$-functions.

\paragraph{\textbf{On-path sampling and stability.}} 
The reparameterization in Eq.~\eqref{eq:reparam} enables gradients to flow through $\mu_i^{(l)}$ and $\sigma_i^{(l)}$. 
To enforce bandwidth constraints, we will regularize these posteriors via KL divergence to an uninformative prior (Sec.~\ref{sec:kl_penalty}).
Crucially, we propagate the \emph{sampled} $z_i^{(l)}$—not its mean $\mu_i^{(l)}$—into the $Q$-functions for value estimation and policy updates.
This \emph{on-path} coupling ensures the bandwidth regularization directly constrains the stochastic representations that drive decision-making, rather than regularizing an auxiliary pathway that bypasses the $Q$-network.
In contrast, \emph{off-path} approaches apply bottlenecks to separate encoder branches whose outputs do not directly inform control, weakening the connection between compression and coordination performance (see Sec.~\ref{sec:ablation_onpath} for empirical validation).

To prevent variance collapse (where $\sigma_i^{(l)} \to 0$ eliminates stochasticity and degrades exploration), we clamp the log-variance:
\begin{equation}
\log \sigma_i^{2(l)} \in [\log \sigma_{\min}^2, \log \sigma_{\max}^2],
\end{equation}
with $\sigma_{\min}{=}0.01$ and $\sigma_{\max}{\in}[1,3]$.
This maintains a minimum level of noise in the messages, encouraging the model to preserve uncertainty rather than collapsing to deterministic point estimates.

\begin{figure*}[t]
\centering
\includegraphics[width=\textwidth ]{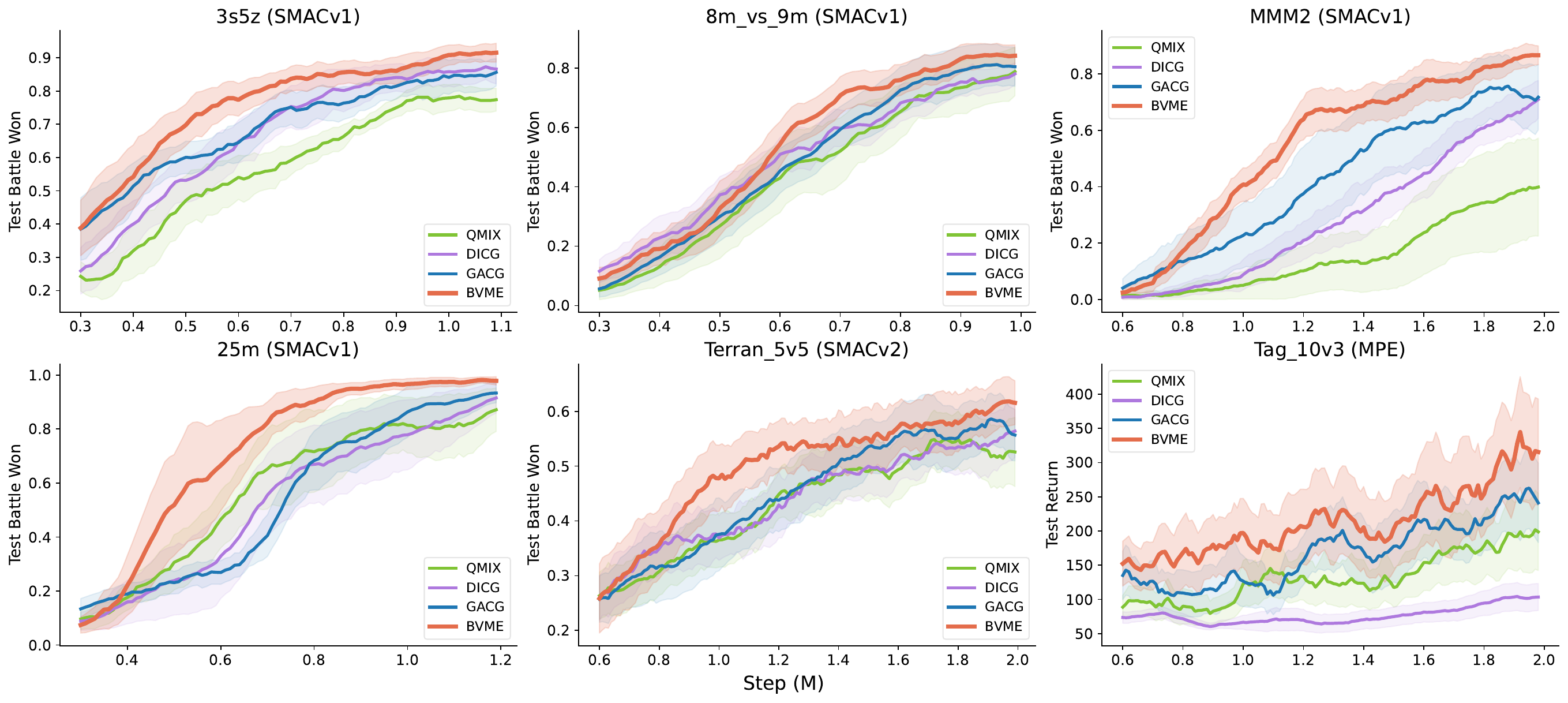}
\caption{\textbf{Overall performance across benchmarks.}
Learning curves on SMACv1 (3s5z, 8m\_vs\_9m, MMM2, 25m), SMACv2 (Terran\_5v5), and MPE Tag at bandwidth ratio $r=0.05$.
BVME consistently outperforms QMIX, DICG, and GACG across all tasks, with shaded regions indicating standard error over 5 seeds.
}
\label{fig:overall_perf}
\end{figure*}

\subsection{Bandwidth Enforcement via KL Divergence}
\label{sec:kl_penalty}

Stochastic sampling alone does not guarantee compression—the model could set $\sigma_i^{(l)} \!\to\! 0$ and encode arbitrarily detailed features in $\mu_i^{(l)}$. 
To enforce meaningful compression, we regularize each agent's message distribution following the information bottleneck principle~\cite{tishby2000information}: limit transmitted information while preserving task-relevant content.

We achieve this by regularizing each agent's posterior $p_i^{(l)}$ toward a shared uninformative prior via KL divergence.
The prior serves as a reference: closer distributions transmit less information, defined as:
\begin{equation}
q(z) = \mathcal{N}(0,\, \sigma_0^2 I),
\label{eq:prior_def}
\end{equation}
a zero-mean Gaussian carrying no task-relevant information—agents transmitting messages sampled from $q(z)$ would communicate nothing useful.
The KL divergence $\mathrm{KL}(p_i^{(l)} \,\|\, q)$ measures how different agent $i$'s message distribution is from this uninformative prior.
By penalizing large KL values, we limit how much agents can deviate from $q(z)$, directly controlling communication capacity.
Note that $q(z)$ has no agent subscript—all agents share the same prior, ensuring identical bandwidth constraints.
When $\sigma_0$ is very small, $q(z)$ approaches a point mass at zero, representing the limit of "transmitting nothing."

\noindent\textbf{Why this enforces bandwidth limits.}
The prior variance $\sigma_0^2$ directly controls the information capacity of each message dimension.
To see this concretely, consider how the KL penalty responds to different encoding strategies.
If $\sigma_0 = 0.1$ (tight capacity), encoding a feature with $\mu_{i,d} = 5$ and $\sigma_{i,d} = 1$ yields a KL contribution of approximately $\frac{25 + 1}{0.01} - 1 \approx 2599$, creating strong pressure to reduce $\mu_{i,d}$ or eliminate this dimension.
Conversely, with $\sigma_0 = 1$ (relaxed capacity), the same configuration yields $\frac{25 + 1}{1} - 1 = 25$, allowing more information transmission.

For a diagonal Gaussian posterior $p_i^{(l)} = \mathcal{N}\!\big(\mu_i^{(l)}, \mathrm{diag}(\sigma_i^{2(l)})\big)$, 
the KL divergence has a closed form:
\begin{equation}
\mathrm{KL}\!\big(p_i^{(l)} \,\|\, q\big)
= \frac{1}{2} \sum_{d=1}^{d_{\mathrm{msg}}}
\left(
\frac{\sigma_{i,d}^2 + \mu_{i,d}^2}{\sigma_0^2}
-1 + \log\frac{\sigma_0^2}{\sigma_{i,d}^2}
\right),
\label{eq:diag_kl}
\end{equation}
where $\mu_{i,d}$ and $\sigma_{i,d}$ index dimension $d$ of agent $i$'s posterior (see Appendix~\ref{sec:kl_derivation} for derivation).
This penalizes both large means ($\mu_{i,d}^2$ term) and variances ($\sigma_{i,d}^2$ term) relative to the prior, encouraging compact representations.
\subsection{Training Objective}
\label{sec:training_objective}

At each timestep $t$, each agent $i$ samples a Bandwidth-constrained message $z_i^t$ via Eq.~\eqref{eq:reparam} applied at the final GNN layer $L$.
The local utility then conditions on this sampled message $
Q_i(\tau_i^t, u_i^t, z_i^t),$
which captures how agent $i$ evaluates action $u_i^t$ given its history $\tau_i^t$ and the compressed message $z_i^t$.
The mixer aggregates $\{Q_i\}_{i=1}^n$ into $Q_{\mathrm{tot}}$ following standard value decomposition (Sec.~\ref{sec:prelim_gacg}).

Let $\mathcal{B}$ denote a replay batch with timestep mask $m_t\!\in\!\{0,1\}$ indicating valid transitions.
The BVME regularization term averages KL divergences across agents and timesteps:
\begin{equation}
\mathcal{L}_{\text{BVME}}
= \frac{\lambda_{\mathrm{KL}}}{n\,T_{\mathcal{B}}}
\sum_{t \in \mathcal{B}} m_t \sum_{i=1}^{n}
\mathrm{KL}\!\big(p_i^{(L,t)} \,\|\, q\big),
\label{eq:bvme_loss}
\end{equation}
where $T_{\mathcal{B}}=\sum_{t\in \mathcal{B}} m_t$ counts valid timesteps and $p_i^{(L,t)}$ denotes agent $i$'s posterior at layer $L$ and time $t$. 
The KL weight $\lambda_{\mathrm{KL}}$ scales the regularization pressure, trading off compression against coordination performance.
Together with the prior scale $\sigma_0$ (Eq.~\ref{eq:prior_def}), these hyperparameters provide complementary control over bandwidth: $\sigma_0$ sets the baseline capacity per dimension, while $\lambda_{\mathrm{KL}}$ determines how strongly to enforce this constraint. While BVME can be applied at any GNN layer, we apply it only at the final layer $L$ whose outputs feed the $Q$-functions, as this directly regularizes the representations used for control.

The full training objective combines the BVME regularization with the GACG backbone:
\begin{equation}
\mathcal{L}
= \underbrace{\mathcal{L}_{\mathrm{TD}} + \lambda_g\,\mathcal{L}_g}_{\text{GACG objective}}
\;+\; \mathcal{L}_{\text{BVME}}.
\label{eq:full_loss}
\end{equation}
During training, we sample stochastic messages via Eq.~\eqref{eq:reparam} to enable exploration and ensure the KL penalty shapes the representations.
At evaluation, we use the deterministic mean $z_i = \mu_i^{(L)}$ to reduce variance and improve stability.

\section{Experiments}
We design our experiments to answer the following key questions:  
(1) How does BVME perform compared to strong MARL baselines across diverse benchmarks?  
(2) How sensitive is performance to the bandwidth ratio $r = d_{\mathrm{msg}} / d_{\mathrm{obs}}$?  
(3) What is the effect of feeding \emph{sampled} variational messages into the $Q$-network (on-path) versus using only the mean (off-path)?  
(4) Is BVME particularly beneficial when combined with sparse-graph backbones (e.g., GACG), where fewer edges make message quality more critical?  
(5) How do the regularization strength $\lambda_{\mathrm{KL}}$ and prior scale $\sigma_0$ jointly affect performance?  

All experiments are conducted on three cooperative MARL benchmarks with results averaged over 5 random seeds. 
\textbf{SMACv1}~\cite{DBLP:conf/atal/SamvelyanRWFNRH19} provides standardized StarCraft II micromanagement scenarios, serving as a widely used benchmark for value-decomposition and graph-based MARL. 
\textbf{SMACv2}~\cite{DBLP:conf/nips/EllisCMSSMFW23} increases stochasticity by randomizing start positions and unit types, testing robustness and generalization. 
\textbf{Tag}~\cite{DBLP:conf/nips/LoweWTHAM17} emphasizes communication under partial observability in a predator-prey scenario.
See Appendix~\ref{sec:imp_details} for full specifications.


\begin{figure}[t]
\centering
\includegraphics[width=\columnwidth]{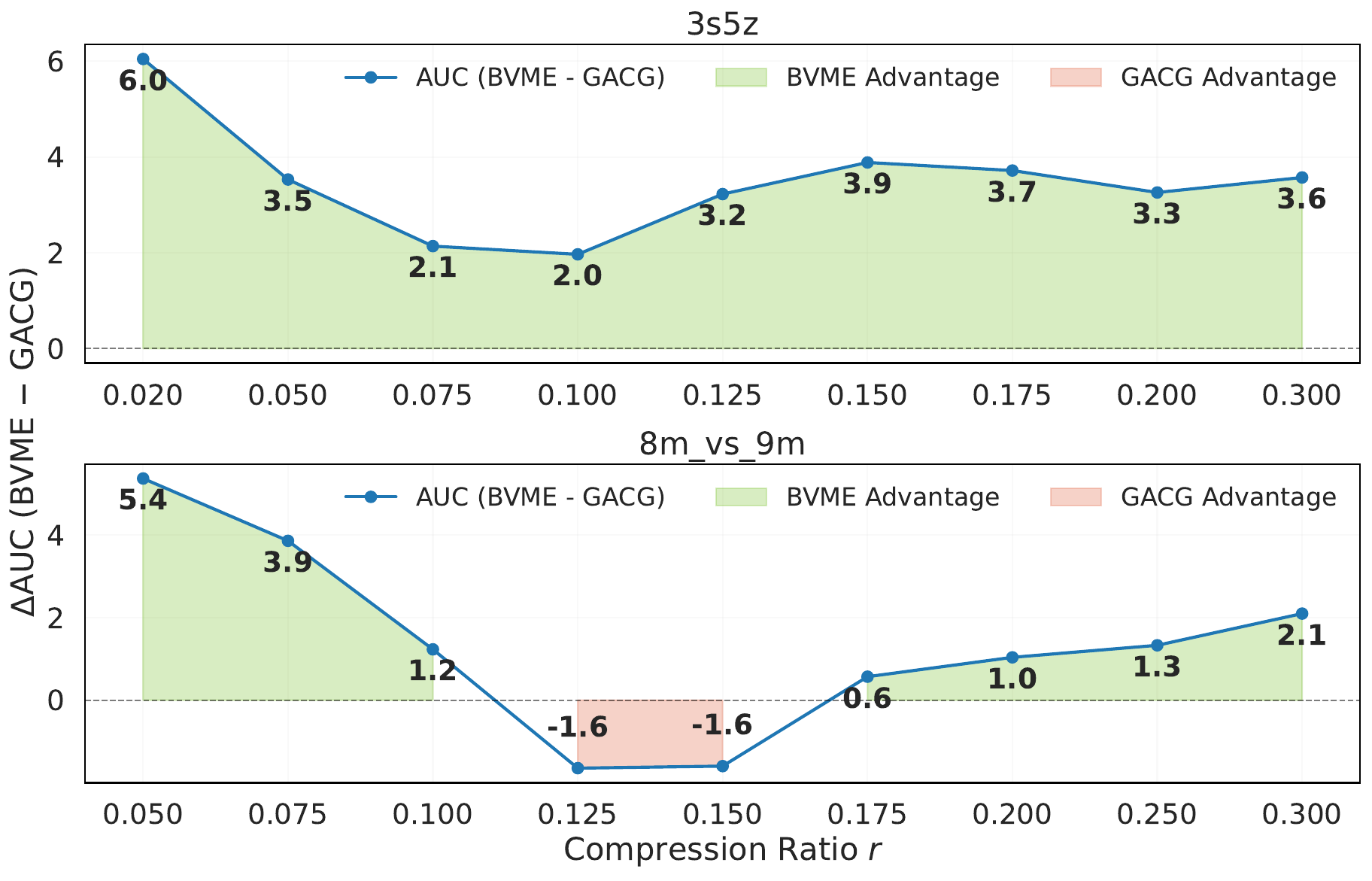}
\caption{
\textbf{U-shaped performance curve.}
Relative cumulative performance $\Delta\mathrm{AUC}(r)=\mathrm{AUC}_{\text{BVME}}-\mathrm{AUC}_{\text{GACG}}$ across compression ratios. 
Green regions indicate BVME advantage; orange regions indicate GACG advantage. 
BVME excels at low ($\le 0.05$) and high ($\ge 0.15$) bandwidths, with GACG competitive only in the mid-range ($0.10$--$0.15$).
}
\label{fig:AUC}
\end{figure}

\subsection{Overall Performance}
\label{sec:overall}

We compare BVME against representative cooperative MARL baselines: 
\begin{itemize}
    \item \textbf{QMIX}~\cite{DBLP:conf/icml/RashidSWFFW18}: value decomposition with centralized training but no graph-structured communication.
    \item \textbf{DICG}~\cite{DBLP:conf/atal/LiGMAK21}: fully connected coordination graph with attention-based message weighting.
    \item \textbf{GACG}~\cite{DBLP:conf/ijcai/00030X24}: state-of-the-art sparse coordination graph learning with group-aware structure discovery.
    \item \textbf{BVME (ours)}: variational Bandwidth-constrained message encoding applied to the GACG backbone.
\end{itemize}

\noindent\textbf{Results.} 
Fig.~\ref{fig:overall_perf} shows learning curves across all benchmarks.
BVME consistently outperforms baselines under bandwidth constraints ($r=0.05$): on small-scale maps (3s5z, 8m\_vs\_9m), BVME achieves 5-10\% higher final win rates than GACG, while on larger scenarios (25m, MMM2), it demonstrates both faster convergence and superior asymptotic performance.
On SMACv2 (Terran\_5v5), BVME maintains a clear advantage over both GACG and DICG despite increased stochasticity, demonstrating robustness to harder, randomized environments. 
On MPE Tag, BVME improves coordination in asymmetric predator-prey interactions, highlighting adaptability beyond StarCraft domains.
These results establish BVME as an effective mechanism for multi-agent coordination under stringent bandwidth budgets, with benefits most pronounced on sparse graphs where message quality critically impacts performance.

\begin{figure}[t]
\centering
\includegraphics[width=\columnwidth]{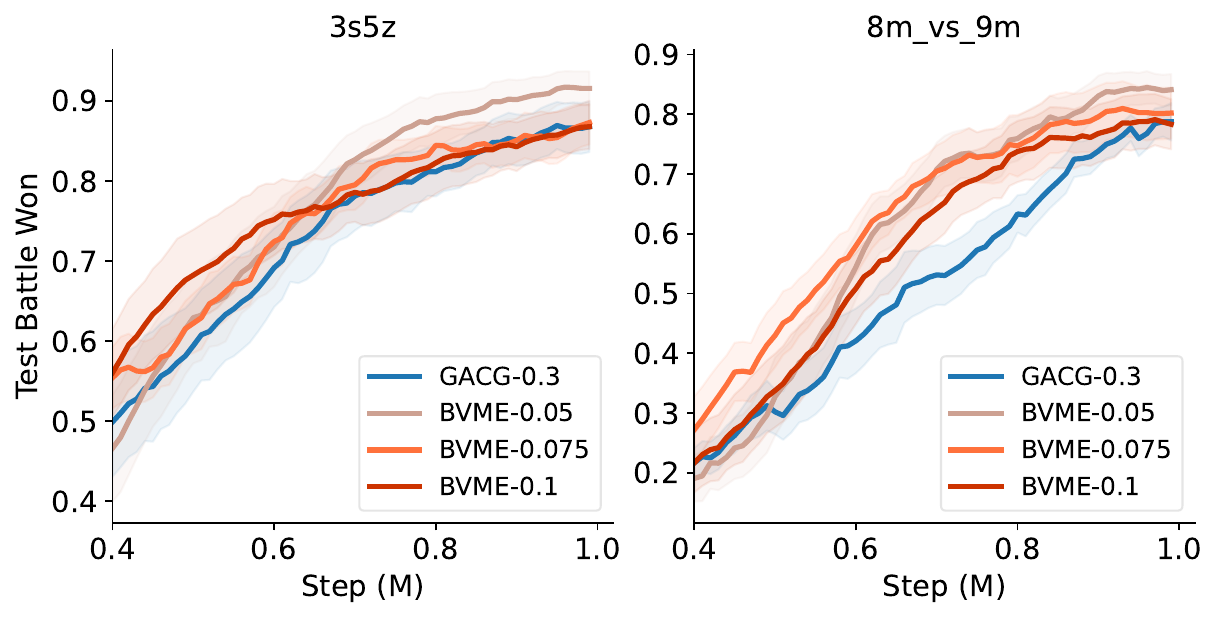}
\caption{\textbf{Extreme compression comparison.} GACG at full budget 
($r{=}0.30$) versus BVME at tighter ratios ($r{=}0.05,0.075,0.10$) on 
\texttt{3s5z} and \texttt{8m\_vs\_9m}. Even with 67–83\% fewer message 
dimensions, BVME achieves comparable or superior win rates.}
\label{fig:Ab-GACG-0.3}
\end{figure}

\begin{table}[t]
\centering
\begin{tabular}{l l c c c c}
\toprule
Map & Method & $r$ & \makecell{Msg Red.\\vs.\ 0.30} & AUC & Win \\
\midrule
\multirow{4}{*}{\texttt{3s5z}} 
& GACG-0.30  & 0.30 & --     & 0.5201 & 0.8677 \\
& BVME-0.05  & 0.05 & 83.3\% & 0.5517 & 0.9156 \\
& BVME-0.075 & 0.075& 75.0\% & 0.5427 & 0.8729 \\
& BVME-0.10  & 0.10 & 66.7\% & 0.5570 & 0.8677 \\
\midrule
\multirow{4}{*}{\texttt{8m\_vs\_9m}} 
& GACG-0.30  & 0.30 & --     & 0.3273 & 0.7875 \\
& BVME-0.05  & 0.05 & 83.3\% & 0.3791 & 0.8411 \\
& BVME-0.075 & 0.075& 75.0\% & 0.4005 & 0.8021 \\
& BVME-0.10  & 0.10 & 66.7\% & 0.3621 & 0.7833 \\
\bottomrule
\end{tabular}
\caption{Comparison of GACG at full budget ($r=0.30$) with BVME at reduced ratios on SMAC maps. 
``Msg Red. vs.\ 0.30'' denotes the relative message dimensionality reduction compared to the $r=0.30$ baseline, computed as $(0.30-r)/0.30$. 
BVME achieves comparable or superior performance while reducing communication by up to 83\%.}
\label{tab:bvme_reduction}
\end{table}



\subsection{Compression Ratio Sensitivity}
\label{sec:ablation_ratio}

The bandwidth ratio $r = d_{\mathrm{msg}} / d_{\mathrm{obs}}$ directly controls message dimensionality. We ask: \textbf{to what extent can BVME preserve coordination effectiveness under aggressive compression, and how does this compare to deterministic compression (GACG)?}

\paragraph{Setup.}
We vary $r$ on \texttt{3s5z} from  $0.02\sim0.30$ and on \texttt{8m\_vs\_9m} from $0.05\sim0.30$, covering extreme bottlenecks (2–5\%) to relaxed budgets (30\%). For each setting, we measure the area under the learning curve (AUC) and final test win rate over 5 seeds. To visualize relative performance, we compute
$
  \Delta \mathrm{AUC}(r) \;=\; \mathrm{AUC}_{\text{BVME}}(r) - \mathrm{AUC}_{\text{GACG}}(r),
$
where positive values indicate BVME achieves stronger cumulative learning than GACG at the same budget (Fig.~\ref{fig:AUC}). We further contrast BVME at low ratios ($r{=}0.05,0.075,0.10$) with GACG at full budget ($r{=}0.30$) in Fig.~\ref{fig:Ab-GACG-0.3} and Tab.~\ref{tab:bvme_reduction}.

\paragraph{Results.}
Fig.~\ref{fig:AUC} reveals a consistent \textbf{U-shaped advantage curve}. At very low $r$ ($\le 0.05$), BVME substantially outperforms GACG by preserving task-critical features through KL-regularized stochastic encoding: on \texttt{3s5z}, $\Delta \mathrm{AUC} = 6.0$ at $r=0.02$ and $3.5$ at $r=0.05$; on \texttt{8m\_vs\_9m}, gains reach $5.4$ and $3.9$ respectively. In the mid-range ($r \approx 0.10$--$0.15$), the gap narrows. On \texttt{8m\_vs\_9m}, GACG even shows slight advantages ($\Delta \mathrm{AUC} = -1.6$ at $r=0.125,0.15$), likely because deterministic projection suffices at these ratios while BVME's KL penalty introduces regularization overhead. However, this regime is narrow: as $r$ increases beyond 0.15, BVME recovers its advantage. At higher budgets ($r \ge 0.15$), BVME again outperforms GACG as KL regularization filters nuisance dimensions and prevents overfitting. On \texttt{3s5z}, $\Delta \mathrm{AUC}$ remains positive across all $r$ values, indicating consistent benefits.

\noindent\textbf{Extreme compression.}
Tab.~\ref{tab:bvme_reduction} quantifies a striking result: BVME at $r=0.05$ (83\% message reduction vs.\ GACG-0.30) achieves \emph{higher} win rates on both maps—0.916 vs.\ 0.868 on \texttt{3s5z} and 0.841 vs.\ 0.788 on \texttt{8m\_vs\_9m}—while also improving AUC. This demonstrates that variational encoding not only preserves coordination under severe bandwidth constraints but can actively improve performance by forcing compact, task-relevant representations.

These findings establish BVME's robustness across bandwidth regimes: substantial gains under tight constraints, competitive performance in mid-ranges, and denoising benefits at high capacity. To validate the underlying mechanisms, we next isolate the role of on-path coupling, test alternative graph backbones, and sweep the regularization hyperparameters $\lambda_{\mathrm{KL}}$ and $\sigma_0$.

\begin{figure}[t]
\centering
\includegraphics[width=\columnwidth]{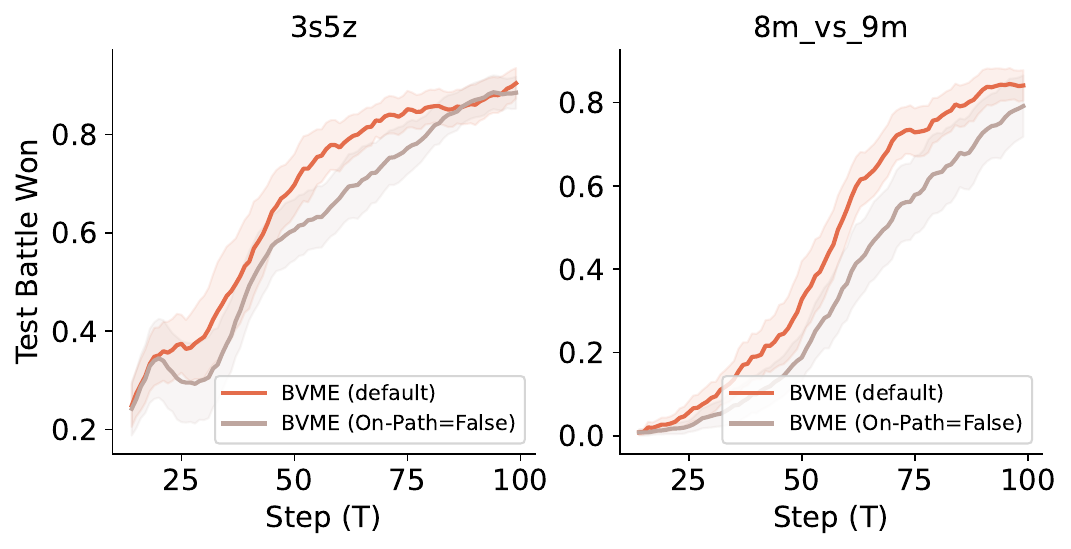}
\caption{
\textbf{On-path coupling is essential for effective bandwidth control.}
Learning curves comparing on-path (stochastic messages $z_i = \mu_i + \sigma_i \odot \varepsilon$ fed to $Q$-network) versus off-path (deterministic mean $z_i = \mu_i$ fed to $Q$-network) at $r=0.05$.
On-path achieves 6-8\% higher win rates by ensuring the KL penalty directly constrains decision-making representations.
}
\label{fig:Ab-OnPath}
\end{figure}

\begin{figure}[t]
\centering
\includegraphics[width=\columnwidth]{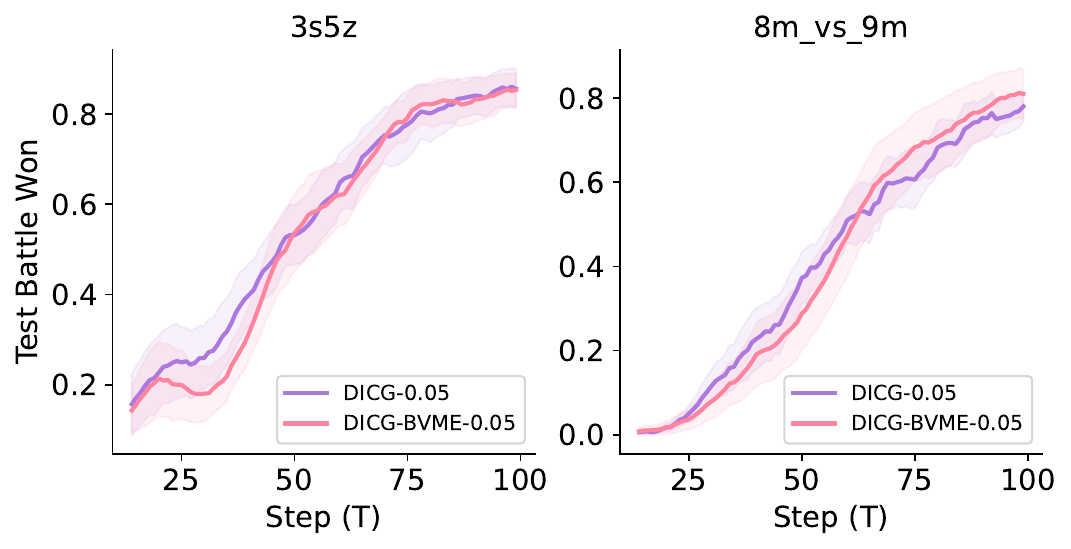}
\caption{BVME provides modest gains on dense-graph backbones.
Learning curves comparing DICG (fully connected) with and without BVME at $r=0.05$. 
Final win rates are nearly identical on \texttt{3s5z} (0.86 vs.\ 0.86) with only a 2\% improvement on \texttt{8m\_vs\_9m} (0.82 vs.\ 0.80), contrasting with 5-10\% gains on sparse GACG.
}
\label{fig:Ab-DICG}
\end{figure}

\subsection{On-Path vs.\ Off-Path Coupling}
\label{sec:ablation_onpath}

A core design choice in BVME is \emph{on-path} coupling: using the sampled stochastic messages $z_i^{(l)}$ (Eq.~\ref{eq:reparam}) directly for $Q$-value estimation, ensuring the KL penalty constrains the representations that drive decisions. 
We validate this choice by comparing on-path training against an \emph{off-path} variant that applies the same KL regularization but feeds only the deterministic mean $\mu_i^{(l)}$ into the $Q$-network.

\paragraph{Setup.}
Both variants use identical architectures and hyperparameters at bandwidth ratio $r{=}0.05$ on \texttt{3s5z} and \texttt{8m\_vs\_9m}, and deterministic messages ($z_i = \mu_i$) at evaluation for stability, differing only in how messages are used during training:
\begin{itemize}
\item \textbf{On-path:} Samples stochastic messages $z_i^{(l)} = \mu_i^{(l)} + \sigma_i^{(l)} \odot \varepsilon$ and feeds them directly to the $Q$-network, ensuring the KL penalty constrains the representations used for decisions.

\item \textbf{Off-path:} Computes the same KL divergence $\mathrm{KL}(p_i^{(l)} \| q)$ using $(\mu_i^{(l)}, \sigma_i^{(l)})$ but forwards only the deterministic mean $z_i^{(l)} = \mu_i^{(l)}$ for value estimation, decoupling regularization from the decision pathway.
\end{itemize}

\paragraph{Results.}
Fig.~\ref{fig:Ab-OnPath} shows on-path coupling achieves both faster convergence and higher asymptotic performance.
On \texttt{3s5z}, on-path reaches 0.88 win rate by 0.8M steps while off-path plateaus near 0.82; on \texttt{8m\_vs\_9m}, the gap widens further with on-path achieving 0.80 vs.\ 0.72 for off-path.

This gap arises because on-path training enforces consistency between the regularized representations and the representations used for control. 
The KL penalty directly shapes the stochastic messages that inform $Q$-values, creating tight coupling between compression and coordination.
In contrast, off-path decouples this link: the encoder learns $(\mu_i, \sigma_i)$ under KL pressure, but since only $\mu_i$ drives decisions, the model can collapse $\sigma_i \to 0$ to minimize the penalty while encoding detailed information in $\mu_i$—effectively bypassing the bandwidth constraint.

These results confirm that on-path coupling is essential for BVME's effectiveness, ensuring the variational bottleneck meaningfully constrains the information used for coordination rather than just an auxiliary pathway.

\subsection{Dense vs.\ Sparse Graph Backbones}
\label{sec:ablation_dicg}

To test whether BVME's benefits generalize beyond sparse structures, we apply it to DICG—a fully connected backbone where all agents can communicate.

\paragraph{Setup.}
We compare \texttt{DICG-0.05} against \texttt{DICG-BVME-0.05} (with variational message encoding) at bandwidth ratio $r{=}0.05$ on \texttt{3s5z} and \texttt{8m\_vs\_9m}, using identical hyperparameters.

\paragraph{Results.}
Fig.~\ref{fig:Ab-DICG} shows BVME provides only modest improvements on dense graphs: on \texttt{3s5z}, final win rates are nearly identical (0.86 for both), while on \texttt{8m\_vs\_9m}, BVME yields only 2\% gain (0.82 vs.\ 0.80), contrasting sharply with 5-10\% improvements on sparse GACG (Sec.~\ref{sec:overall}).
This aligns with our hypothesis: in fully connected topologies, agents aggregate complementary information from all teammates, providing rich information flow even under compression. 
BVME's variational encoding offers limited benefit because agents need not rely on a few critical edges.
Conversely, in sparse graphs, each edge carries more coordination responsibility, making message quality crucial.
These results validate BVME's design: it excels when communication channels are scarce, as KL-regularized encoding prioritizes task-critical information on limited edges, while gains are marginal on dense topologies where connectivity is already maximal.

\begin{figure}[t]
\centering
\includegraphics[width=\columnwidth]{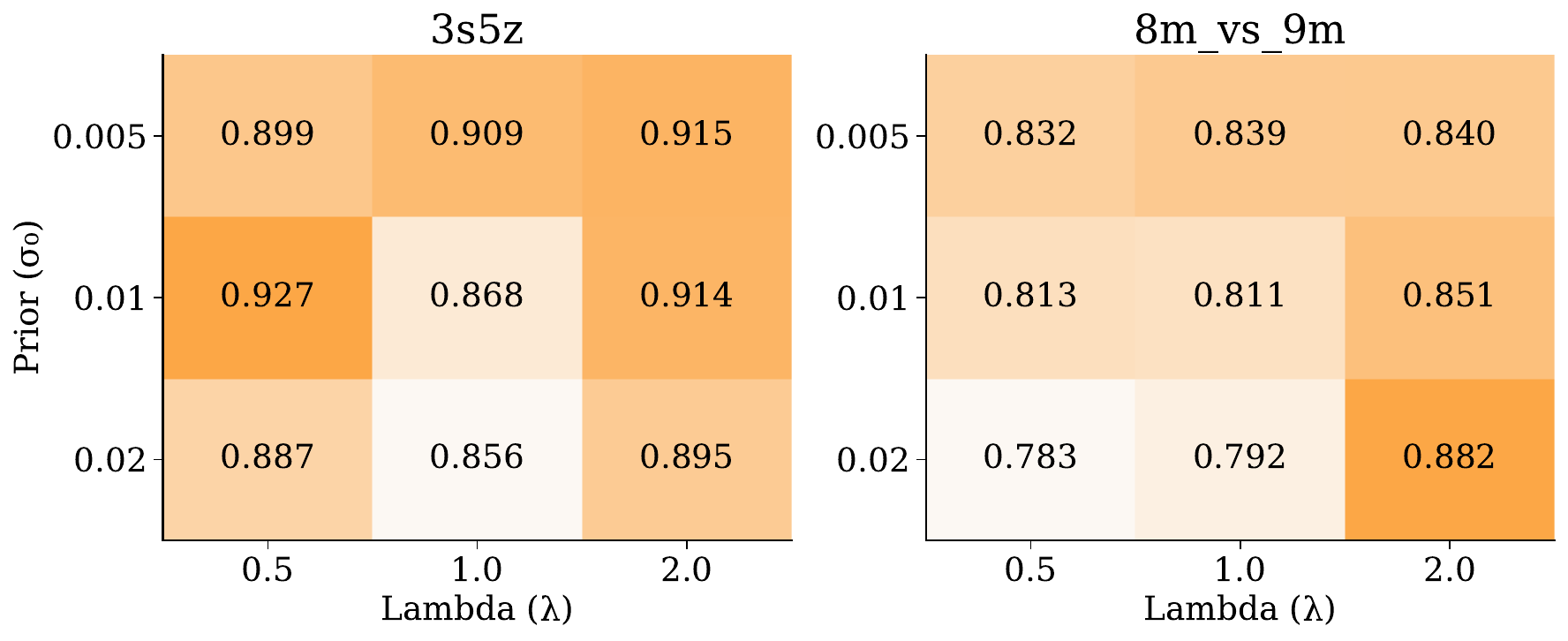}
\caption{
Mean test win rates across KL weight $\lambda_{\mathrm{KL}}$ and prior scale $\sigma_0$ at $r{=}0.1$. 
Optimal settings differ by task: \texttt{3s5z} peaks at $(\lambda_{\mathrm{KL}}{=}0.5, \sigma_0{=}0.01)$ achieving 0.927, while \texttt{8m\_vs\_9m} prefers $(\lambda_{\mathrm{KL}}{=}2.0, \sigma_0{=}0.02)$ achieving 0.882.
}
\label{fig:Ab-StrengthPrior}
\end{figure}

\subsection{Hyperparameter Sensitivity}
\label{sec:ablation_strength_prior}

BVME involves two key hyperparameters that jointly govern compression strength: the KL weight $\lambda_{\mathrm{KL}}$ controls penalty strength, while the prior variance $\sigma_0^2$ sets the baseline capacity.
We investigate how their interaction affects coordination performance.

\paragraph{Setup.}
We perform a $3\times3$ grid search over $\lambda_{\mathrm{KL}} \in \{0.5, 1.0, 2.0\}$ and $\sigma_0 \in \{0.005, 0.01, 0.02\}$, evaluating mean test win rates over 5 seeds on \texttt{3s5z} and \texttt{8m\_vs\_9m}, under fixed bandwidth ratio $r=0.05$.

\paragraph{Results.}
Fig.~\ref{fig:Ab-StrengthPrior} reveals task-dependent sensitivity to regularization strength. 
\textbf{3s5z:} Performance peaks at $(\lambda_{\mathrm{KL}}{=}0.5, \sigma_0{=}0.01)$ with 0.927 win rate, remaining strong across moderate settings but dropping to 0.868 at $(\lambda_{\mathrm{KL}}{=}1.0, \sigma_0{=}0.01)$, suggesting over-regularization.
\textbf{8m\_vs\_9m:} The optimal setting shifts to $(\lambda_{\mathrm{KL}}{=}2.0, \sigma_0{=}0.02)$ achieving 0.882, with performance degrading substantially under weak regularization: $(\lambda_{\mathrm{KL}}{=}0.5, \sigma_0{=}0.02)$ yields only 0.783.
BVME shows reasonable robustness (performance spans 6-10\% across the grid) but benefits from task-specific tuning.
\textbf{Moderate compression pressure extracts useful features, while extremes either fail to eliminate redundancy or remove critical information.}
The differing optimal settings across tasks suggest that coordination complexity determines the ideal regularization trade-off.

\section{Discussion}

\paragraph{Computational Complexity.}
Let $n$ denote the number of agents, $d_{\mathrm{obs}}$ the observation dimension, and $d_{\mathrm{msg}} = r \cdot d_{\mathrm{obs}}$ the message dimension with $r \ll 1$.
BVME adds $\mathcal{O}(n \cdot d_{\mathrm{msg}}^2)$ operations for two encoder MLPs ($\mathrm{Enc}_\mu$, $\mathrm{Enc}_\sigma$), $\mathcal{O}(n \cdot d_{\mathrm{msg}})$ for reparameterized sampling, and $\mathcal{O}(n \cdot d_{\mathrm{msg}})$ for closed-form KL divergence.
Since $d_{\mathrm{msg}}^2 = r^2 \cdot d_{\mathrm{obs}}^2$ with $r \approx 0.05$, overhead is negligible compared to the baseline's $\mathcal{O}(L \cdot n^2 \cdot d_{\mathrm{obs}}^2)$ message-passing cost.
Wall-clock training time increases by $<5\%$ on SMAC; at evaluation, deterministic means ($z_i = \mu_i$) incur zero additional cost.

\paragraph{Relation to Prior Bandwidth-constrained Methods.}

MAGI~\cite{DingMAGI2024} is the most closely related work, both applying information bottleneck principles~\cite{tishby2000information} to multi-agent communication but at different levels and for different objectives.
MAGI applies bottlenecks to \emph{graph structure}, filtering noisy channels to achieve robustness against adversarial attacks via hierarchical dual-MI optimization over node features and edge structures.
BVME applies bottlenecks to \emph{individual messages}: each message is a sample from a learned Gaussian distribution regularized toward an uninformative prior, compressing agent observations into bandwidth-limited representations under hard dimensional constraints ($d_{\mathrm{msg}} \ll d_{\mathrm{obs}}$).

Key differences: \textbf{(1)} BVME provides explicit dimensional capacity control via compression ratio $r$ and prior scale $\sigma_0$ with closed-form KL (Eq.~\ref{eq:diag_kl}), enabling practitioners to directly tune the bandwidth-performance trade-off; MAGI's robustness terms regulate information flow but do not enforce dimensional constraints. \textbf{(2)} BVME uses lightweight diagonal-Gaussian encoding at a single GNN layer ($<$5\% overhead), while MAGI requires nested variational bounds, categorical and Gaussian sampling across multiple layers, and adversarial training. \textbf{(3)} BVME's on-path regularization ensures the KL penalty directly constrains representations used for decision-making (Sec.~\ref{sec:ablation_onpath}), rather than auxiliary pathways.

\section{Conclusion}

We introduced BVME, a variational message encoding framework for Bandwidth-constrained multi-agent coordination.
Hard bandwidth constraints force selective encoding, but deterministic projections lack control over how compression occurs.
BVME addresses this by modeling messages as samples from learned Gaussian posteriors regularized via KL divergence, providing tunable compression control through interpretable hyperparameters. BVME's lightweight design—diagonal Gaussians, closed-form KL regularization, and on-path coupling—integrates seamlessly with existing coordination-graph methods while adding minimal overhead.

Across SMACv1, SMACv2, and MPE benchmarks, BVME achieves comparable or superior performance using 67--83\% fewer message dimensions. Ablations reveal U-shaped bandwidth sensitivity, with BVME excelling at extreme compression ($r \le 0.05$) and benefits most pronounced on sparse graphs where message quality critically impacts coordination.
On-path coupling proves essential, substantially outperforming off-path regularization.



\begin{acks}
  This work is supported by the Australian Research Council under Australian Laureate Fellowships FL190100149.
\end{acks}


\bibliographystyle{ACM-Reference-Format} 
\bibliography{AAMAS26}


\clearpage
\appendix

\section{Closed-Form KL Divergence Derivation}
\label{sec:kl_derivation}

We derive the closed-form expression for the diagonal Gaussian KL divergence used in BVME (Eq.~\ref{eq:diag_kl}).

\paragraph{General multivariate Gaussian KL}
For a $k$-dimensional Gaussian distribution,
\begin{equation}
\label{eq:k-gaussians}
\mathcal{N}(z;\mu,\Sigma)
= \frac{1}{(2\pi)^{k/2}|\Sigma|^{1/2}}
  \exp\!\left(-\tfrac{1}{2} (z-\mu)^\top \Sigma^{-1} (z-\mu)\right),
\end{equation}
the log-density is
\begin{equation}
\label{eq:gauss-log}
\log \mathcal{N}(z;\mu,\Sigma)
= -\tfrac{1}{2}\!\left(k\log(2\pi) + \log|\Sigma| + (z-\mu)^\top \Sigma^{-1} (z-\mu)\right).
\end{equation}

The KL divergence between two Gaussians $\mathbb{P}=\mathcal{N}(\mu_p,\Sigma_p)$ and $\mathbb{Q}=\mathcal{N}(\mu_q,\Sigma_q)$ is:
\begin{equation}
\mathrm{KL}(\mathbb{P}\,\|\,\mathbb{Q})
= \mathbb{E}_{\mathbb{P}}\!\left[\log \mathcal{N}(Z;\mu_p,\Sigma_p) - \log \mathcal{N}(Z;\mu_q,\Sigma_q)\right].
\end{equation}

Using the identities:
\begin{align}
\mathbb{E}_{\mathbb{P}}\!\big[(Z-\mu_p)^\top\Sigma_p^{-1}(Z-\mu_p)\big] &= k, \\
\mathbb{E}_{\mathbb{P}}\!\big[(Z-\mu_q)^\top\Sigma_q^{-1}(Z-\mu_q)\big] &= \mathrm{tr}\!\big(\Sigma_q^{-1}\Sigma_p\big)
+ (\mu_p-\mu_q)^\top \Sigma_q^{-1}(\mu_p-\mu_q),
\end{align}
we obtain the general formula:
\begin{equation}
\label{eq:gauss-kl-general}
\mathrm{KL}(\mathbb{P}\,\|\,\mathbb{Q})
=\frac{1}{2}\!\left(
\mathrm{tr}\!\big(\Sigma_q^{-1}\Sigma_p\big)
+ (\mu_p-\mu_q)^\top \Sigma_q^{-1}(\mu_p-\mu_q)
- k
+ \log\frac{|\Sigma_q|}{|\Sigma_p|}
\right).
\end{equation}

\paragraph{BVME special case: diagonal posterior, isotropic prior.}
In BVME, each agent $i$ has a diagonal Gaussian posterior $p_i(z) = \mathcal{N}(\mu_i, \mathrm{diag}(\sigma_i^2))$ and shares an isotropic prior $q(z) = \mathcal{N}(0, \sigma_0^2 I)$. Substituting into Eq.~\eqref{eq:gauss-kl-general}:
\begin{itemize}
\item $\Sigma_p = \mathrm{diag}(\sigma_{i,1}^2, \ldots, \sigma_{i,k}^2)$, so $|\Sigma_p| = \prod_{d=1}^k \sigma_{i,d}^2$
\item $\Sigma_q = \sigma_0^2 I$, so $|\Sigma_q| = \sigma_0^{2k}$ and $\Sigma_q^{-1} = \sigma_0^{-2} I$
\item $\mu_p = \mu_i$, $\mu_q = 0$
\end{itemize}

The trace term becomes:
\begin{equation}
\mathrm{tr}\!\big(\Sigma_q^{-1}\Sigma_p\big) = \mathrm{tr}\!\left(\sigma_0^{-2} \mathrm{diag}(\sigma_{i,1}^2, \ldots, \sigma_{i,k}^2)\right) = \sum_{d=1}^k \frac{\sigma_{i,d}^2}{\sigma_0^2}.
\end{equation}

The quadratic term becomes:
\begin{equation}
(\mu_p-\mu_q)^\top \Sigma_q^{-1}(\mu_p-\mu_q) = \mu_i^\top (\sigma_0^{-2} I) \mu_i = \sum_{d=1}^k \frac{\mu_{i,d}^2}{\sigma_0^2}.
\end{equation}

The log-determinant term becomes:
\begin{equation}
\log\frac{|\Sigma_q|}{|\Sigma_p|} = \log\frac{\sigma_0^{2k}}{\prod_{d=1}^k \sigma_{i,d}^2} = \sum_{d=1}^k \log\frac{\sigma_0^2}{\sigma_{i,d}^2}.
\end{equation}

Combining these terms yields:
\begin{equation}
\label{eq:diag-kl-derived}
\mathrm{KL}\!\big(p_i(z) \,\|\, q(z)\big)
=
\frac{1}{2}\sum_{d=1}^{k}\left(
\frac{\sigma_{i,d}^2+\mu_{i,d}^2}{\sigma_0^2}
-1
+\log\frac{\sigma_0^2}{\sigma_{i,d}^2}
\right),
\end{equation}
where $k = d_{\mathrm{msg}}$. This is Eq.~\eqref{eq:diag_kl} in the main paper.

\paragraph{Interpretation.}
Each dimension $d$ contributes three terms to the KL penalty:
\begin{itemize}
\item $\frac{\sigma_{i,d}^2}{\sigma_0^2}$: penalizes large posterior variance (diffuse encoding)
\item $\frac{\mu_{i,d}^2}{\sigma_0^2}$: penalizes large posterior mean (high-magnitude features)
\item $\log\frac{\sigma_0^2}{\sigma_{i,d}^2}$: entropy regularization (encourages posterior to match prior scale)
\end{itemize}
Together, these terms encourage compact, low-variance representations centered near zero, effectively enforcing bandwidth constraints.

\paragraph{Normalization.}
To ensure consistent regularization strength across different compression ratios $r$, we normalize the KL penalty by $d_{\mathrm{msg}}$:
\begin{equation}
\mathcal{L}_{\mathrm{KL}} = \frac{\lambda_{\mathrm{KL}}}{n \cdot d_{\mathrm{msg}}} \sum_{i=1}^n \sum_{d=1}^{d_{\mathrm{msg}}} \left(
\frac{\sigma_{i,d}^2+\mu_{i,d}^2}{\sigma_0^2}
-1
+\log\frac{\sigma_0^2}{\sigma_{i,d}^2}
\right),
\end{equation}
where $\lambda_{\mathrm{KL}}$ is the KL weight and $n$ is the number of agents. This ensures the regularization strength remains consistent across different message dimensions.
\section{Implementation Details}
\label{sec:imp_details}

\subsection{Network Architectures}

\paragraph{MLP Encoder.}
The MLP encoder compresses raw observations into initial agent features:
\begin{equation}
h_i^{(0)} = \text{ReLU}(\mathbf{W}_1 o_i + b_1), \quad h_i^{(0)} \in \mathbb{R}^{d_{\text{msg}}}
\end{equation}
where $o_i \in \mathbb{R}^{d_{\text{obs}}}$ is agent $i$'s observation and $d_{\text{msg}} = r \cdot d_{\text{obs}}$ with compression ratio $r$.

\paragraph{GNN Layers.}
We use $L=2$ graph convolution layers following GACG~\cite{DBLP:conf/ijcai/00030X24}. Each layer aggregates neighbor features via the learned adjacency $\tilde{A}$:
\begin{equation}
h_i^{(l+1)} = \text{ReLU}\left(\sum_{j=1}^n \tilde{A}_{ij} \mathbf{W}^{(l)} h_j^{(l)}\right), \quad l = 0, 1
\end{equation}
where $\mathbf{W}^{(l)} \in \mathbb{R}^{d^{(l)} \times d^{(l+1)}}$ are learnable weights. We set $d^{(1)} = d^{(2)} = 128$ for all experiments.

\paragraph{Variational Encoders.}
BVME applies two single-layer MLPs to the final GNN output $h_i^{(L)}$ to parameterize the posterior:
\begin{align}
\mu_i &= \mathbf{W}_\mu h_i^{(L)} + b_\mu, \quad \mu_i \in \mathbb{R}^{d_{\text{msg}}} \\
\log \sigma_i^2 &= \text{clamp}\left(\mathbf{W}_\sigma h_i^{(L)} + b_\sigma, -5, 3\right)
\end{align}
The log-variance is clamped to $[-5, 3]$ for numerical stability, corresponding to standard deviations in $[\exp(-2.5) \approx 0.08, \exp(1.5) \approx 4.48]$.

\paragraph{Agent Network.}
Each agent uses a GRU-based recurrent network to compute $Q$-values from the concatenated input $x_i = [o_i; z_i]$:
\begin{align}
\tilde{h}_i &= \text{ReLU}(\mathbf{W}_1 x_i + b_1) \\
h_i^t &= \text{GRU}(\tilde{h}_i, h_i^{t-1}) \\
Q_i(\tau_i, u_i, z_i) &= \mathbf{W}_2 h_i^t + b_2
\end{align}
where $h_i^t \in \mathbb{R}^{64}$ is the GRU hidden state and $\mathbf{W}_1 \in \mathbb{R}^{64 \times (d_{\text{obs}} + d_{\text{msg}})}$, $\mathbf{W}_2 \in \mathbb{R}^{|U| \times 64}$.

\paragraph{QMIX Mixer.}
Following~\cite{DBLP:conf/icml/RashidSWFFW18}, we use a hypernetwork-generated mixing network:
\begin{equation}
Q_{\text{tot}}(\boldsymbol{\tau}, \mathbf{u}, s) = f_{\text{mix}}(Q_1, \ldots, Q_n; s)
\end{equation}
where $f_{\text{mix}}$ has monotonic weights $\partial Q_{\text{tot}} / \partial Q_i \geq 0$ to satisfy IGM.

\subsection{Training Hyperparameters}

\begin{table}[h]
\centering
\small
\begin{tabular}{lc}
\toprule
\textbf{Parameter} & \textbf{Value} \\
\midrule
Optimizer & RMSprop \\
Learning rate & $5 \times 10^{-4}$ \\
Batch size & 32 \\
Replay buffer size & 5000 episodes \\
Target network update & Every 200 episodes \\
$\gamma$ (discount factor) & 0.99 \\
$\epsilon$-greedy decay & Linear: $1.0 \to 0.05$ over 50k steps \\
Gradient clipping & 10.0 \\
\midrule
\multicolumn{2}{c}{\textit{BVME-specific}} \\
\midrule
Compression ratio $r$ & \{0.02, 0.05, 0.075, 0.10, \ldots, 0.30\} \\
KL weight $\lambda_{\text{KL}}$ & \{0.5, 1.0, 2.0\} \\
Prior scale $\sigma_0$ & \{0.005, 0.01, 0.02\} \\
KL warmup steps & 0 (no warmup) \\
Variance clamp & $\log \sigma^2 \in [-5, 3]$ \\
Training mode & On-path sampling \\
Evaluation mode & Deterministic ($z_i = \mu_i$) \\
\bottomrule
\end{tabular}
\caption{Training hyperparameters for all experiments. BVME uses identical base settings to GACG except for the variational encoding parameters.}
\label{tab:hyperparameters}
\end{table}

\subsection{Pseudocode}

\begin{algorithm}[h]
\caption{BVME Forward Pass (Training)}
\label{alg:bvme_forward}
\begin{algorithmic}[1]
\Require Observations $\{o_i\}_{i=1}^n$, learned adjacency $\tilde{A}$
\Ensure Agent outputs $\{Q_i\}_{i=1}^n$, KL penalty
\State $h_i^{(0)} \gets \text{MLPEncoder}(o_i)$ for all $i$ \Comment{Compress to $d_{\text{msg}}$}
\For{$l = 0$ to $L-1$}
    \State $h_i^{(l+1)} \gets \text{GNN}^{(l)}(h^{(l)}, \tilde{A})$ \Comment{Message passing}
\EndFor
\State $\mu_i, \log\sigma_i^2 \gets \text{Enc}_\mu(h_i^{(L)}), \text{Enc}_\sigma(h_i^{(L)})$ \Comment{Variational encoding}
\State $\varepsilon \sim \mathcal{N}(0, I)$
\State $z_i \gets \mu_i + \sigma_i \odot \varepsilon$ \Comment{Reparameterization}
\State $Q_i \gets \text{AgentNet}([\tau_i, u_i, z_i])$ \Comment{Compute $Q$-values}
\State $\mathcal{L}_{\text{KL}} \gets \frac{1}{n} \sum_{i=1}^n \text{KL}\big(p_i(z) \,\|\, q(z)\big)$ \Comment{Eq.~\ref{eq:diag_kl}}
\State \Return $\{Q_i\}_{i=1}^n$, $\mathcal{L}_{\text{KL}}$
\end{algorithmic}
\end{algorithm}

\begin{algorithm}[h]
\caption{BVME Training Step}
\label{alg:bvme_training}
\begin{algorithmic}[1]
\Require Replay buffer $\mathcal{D}$, batch size $B$, KL weight $\lambda_{\text{KL}}$
\State Sample batch $\{(\tau^{(b)}, \mathbf{u}^{(b)}, r^{(b)}, s^{(b)})\}_{b=1}^B \sim \mathcal{D}$
\State $\{Q_i^{(b)}\}, \mathcal{L}_{\text{KL}}^{(b)} \gets \text{BVME-Forward}(\tau^{(b)})$ \Comment{Alg.~\ref{alg:bvme_forward}}
\State $Q_{\text{tot}}^{(b)} \gets \text{QMIX}(Q_1^{(b)}, \ldots, Q_n^{(b)}, s^{(b)})$
\State $y^{(b)} \gets r^{(b)} + \gamma \max_{\mathbf{u}'} Q_{\text{tot}}^{\text{target}}(\tau'^{(b)}, \mathbf{u}', s'^{(b)})$
\State $\mathcal{L}_{\text{TD}} \gets \frac{1}{B} \sum_{b=1}^B (Q_{\text{tot}}^{(b)} - y^{(b)})^2$
\State $\mathcal{L} \gets \mathcal{L}_{\text{TD}} + \lambda_{\text{KL}} \mathcal{L}_{\text{KL}}$
\State Update parameters via RMSprop on $\nabla \mathcal{L}$
\end{algorithmic}
\end{algorithm}

\subsection{Computational Complexity}

Per training step with batch size $B$, $n$ agents, and $T$ timesteps:
\begin{itemize}
\item \textbf{MLP encoding:} $\mathcal{O}(B \cdot T \cdot n \cdot d_{\text{obs}} \cdot d_{\text{msg}})$
\item \textbf{GNN message passing:} $\mathcal{O}(B \cdot T \cdot L \cdot n^2 \cdot d_{\text{msg}}^2)$ assuming dense adjacency
\item \textbf{Variational encoding:} $\mathcal{O}(B \cdot T \cdot n \cdot d_{\text{msg}}^2)$ (two MLPs: $\text{Enc}_\mu$, $\text{Enc}_\sigma$)
\item \textbf{Reparameterization:} $\mathcal{O}(B \cdot T \cdot n \cdot d_{\text{msg}})$
\item \textbf{KL divergence:} $\mathcal{O}(B \cdot T \cdot n \cdot d_{\text{msg}})$ (closed-form)
\item \textbf{Agent network:} $\mathcal{O}(B \cdot T \cdot n \cdot (d_{\text{obs}} + d_{\text{msg}}) \cdot d_{\text{hidden}})$
\end{itemize}
Since $d_{\text{msg}} = r \cdot d_{\text{obs}}$ with $r \ll 1$ (typically 0.05), the variational encoding overhead is negligible compared to the GNN's $\mathcal{O}(L \cdot n^2 \cdot d_{\text{obs}}^2)$ baseline cost.

\subsection{Environment Details}

\paragraph{SMACv1.} 
We use StarCraft II version 4.10 with default game settings at difficulty 7. 
SMACv1 provides standardized micromanagement scenarios with both homogeneous teams (e.g., 25m with 25 Marines) and heterogeneous compositions (e.g., MMM2 with Marines, Marauders, and Medivacs).
We select these maps to cover small-scale coordination (3s5z with 8 agents), asymmetric combat (8m\_vs\_9m), mixed unit types (MMM2), and large-scale homogeneous teams (25m).

\paragraph{SMACv2.}
Terran\_5v5 increases stochasticity by randomizing start positions (sampled from predefined zones) and unit types (drawn from Terran unit pool) while correcting sight range bugs from SMACv1.
Observation dimension: 82. 
This map tests robustness to distribution shift and adaptability to varying team compositions.

\paragraph{MPE Tag.}
The predator-prey scenario tasks 10 slower predators (max speed 1.0) with capturing 3 faster prey (max speed 1.3) in a $1 \times 1$ continuous 2D world with 2 large circular obstacles (radius 0.15).
Predators must coordinate to surround prey, as individual agents cannot catch them alone.
Observation dimension: 55 (includes relative positions and velocities of all agents and obstacles within a limited sensing radius).
We choose Tag to emphasize communication-critical coordination under partial observability, where agents must share positional information to execute successful encirclement strategies.
\subsection{Decentralized Execution of GACG}

While GACG learns coordination graphs centrally during training, execution can be decentralized in three ways:

\textbf{(1) Dynamic local computation.}
Each agent $i$ independently computes its outgoing edge weights via $\mu_{A}^{t}[i,:] = f_\theta(o_i^t, \{z_j^{t-1} : j \in \mathcal{N}_i\})$, where $\mathcal{N}_i$ is agent $i$'s neighborhood and $f_\theta$ is a learned network~\cite{DBLP:conf/ijcai/00030X24}.
Agent $i$ selects top-$k$ neighbors based on $\mu_{A}^{t}[i,:]$ for message transmission.

\textbf{(2) Static sampled graph.}
Sample a fixed graph once from the learned distribution $\mathcal{N}(\mu_A, \Sigma_A)$ at the end of training and use it throughout deployment.

\textbf{(3) Mean graph.}
Use the mean adjacency $\mu_A$ directly as a fixed graph, avoiding sampling entirely.

BVME is compatible with all three approaches: agents compute compressed messages $z_i = \mu_i$ locally via Eq.~\eqref{eq:reparam} and transmit only to neighbors determined by the chosen graph strategy, requiring no central coordinator.

\end{document}